\documentclass[lettersize,journal]{IEEEtran}
\usepackage{amsmath,amsfonts}
\usepackage{algorithm}
\usepackage{array}

\usepackage{subcaption}
\usepackage{textcomp}
\usepackage{stfloats}
\usepackage{url}
\usepackage{verbatim}
\usepackage{graphicx}
\usepackage{cite}
\usepackage{ifthen}
\usepackage{times}  
\usepackage{helvet}  
\usepackage{courier}  

\usepackage{algpseudocode}

\usepackage{newfloat}
\usepackage{listings}

\usepackage{amssymb}
\usepackage{bbm}
\usepackage{booktabs}
\usepackage{multirow}
\usepackage{makecell}
\usepackage{tabulary}
\usepackage[table,xcdraw]{xcolor}

\newboolean{colored}
\setboolean{colored}{true} 

\newcommand{\revisecolor}[1]{#1}

\begin{document}
\title{Exploring the Coordination of Frequency and Attention in Masked Image Modeling}

\author{Jie~Gui,~\IEEEmembership{Senior Member,~IEEE,}
        Tuo~Chen,
        Minjing~Dong,
        Zhengqi~Liu,
        Hao~Luo,
        James Tin-Yau Kwok,~\IEEEmembership{Fellow,~IEEE,}
        Yuan Yan Tang,~\IEEEmembership{Life Fellow,~IEEE}
\thanks{J. Gui is with the School of Cyber Science and Engineering, Southeast University and with Purple Mountain Laboratories, Nanjing 210000, China (e-mail: guijie@seu.edu.cn).}
\thanks{T. Chen is with the School of Cyber Science and Engineering, Southeast University (e-mail: tchen@seu.edu.cn).}
\thanks{Minjing Dong is with 
Department of Computer Science, City University of Hong Kong (e-mail: minjdong@cityu.edu.hk).}
\thanks{Z. Liu is with the School of Cyber Science and Engineering, Southeast University (e-mail: lzq\_oscar@seu.edu.cn).}
\thanks{H. Luo is with Alibaba Group, Hangzhou 310052, China (e-mail: haoluocsc@zju.edu.cn).}
\thanks{J. Kwok is with the Department of Computer Science and Engineering, The Hong Kong University of Science and Technology, Hong Kong 999077, China. (e-mail: jamesk@cse.ust.hk).}
\thanks{Y. Y. Tang is with Zhuhai UM Science and Technology Research Institute, and also with Faculty of Science \& Technology at University of Macau (e-mail: yytang@um.edu.mo).}
}

\markboth{Journal of \LaTeX\ Class Files,~Vol.~14, No.~8, August~2021}%
{Shell \MakeLowercase{\textit{et al.}}: A Sample Article Using IEEEtran.cls for IEEE Journals}


\maketitle

\begin{abstract}

    Recently, masked image modeling (MIM), which learns visual representations by reconstructing the masked patches of an image, has dominated self-supervised learning in computer vision.
    However, the pre-training of MIM always takes massive time due to the large-scale data and large-size backbones. We mainly attribute it to the random patch masking in previous MIM works, which fails to leverage the crucial semantic information for effective visual representation learning.
    To tackle this issue, we propose the Frequency \& Attention-driven Masking and Throwing Strategy (FAMT), which can extract semantic patches and reduce the number of training patches to boost model performance and training efficiency simultaneously. Specifically, FAMT utilizes the self-attention mechanism to extract semantic information from the image for masking during training in an unsupervised manner.
    However, attention alone could sometimes focus on inappropriate areas regarding the semantic information. Thus, we are motivated to incorporate the information from the frequency domain into the self-attention mechanism to derive the sampling weights for masking, which captures semantic patches for visual representation learning. 
    Furthermore, we introduce a patch throwing strategy based on the derived sampling weights to reduce the training cost. FAMT can be seamlessly integrated as a plug-and-play module and surpasses previous works, \emph{e.g.} reducing the training phase time by nearly $50\%$ and improving the linear probing accuracy of MAE by $1.3\% \sim 3.9\%$ across various datasets, including CIFAR-10/100, Tiny ImageNet, and ImageNet-1K. FAMT also demonstrates superior performance in downstream detection and segmentation tasks.

\end{abstract}

\section{Introduction}

Self-supervised learning (SSL) has gained significant attention in the computer vision field due to its ability to learn the representation of many unlabeled images. A technique called masked image modeling (MIM), inspired by masked language modeling (MLM) in language domain~\cite{gpt3,bert}, has demonstrated its potential in various vision tasks such as classification, object detection, and segmentation~\cite{coco,lvis}. Several state-of-the-art methods have been developed in the past year, including BEiT~\cite{beit}, MAE~\cite{mae}, SimMIM~\cite{SimMIM}, and MaskFeat~\cite{MaskFeat}, which recover masked patches of images to provide the self-supervised signal. With an appropriate masking strategy, MIM can learn general visual representations effectively.
 
\begin{figure}[t]
    \centering
    \begin{subfigure}{0.11\textwidth}
        \centering
        \includegraphics[width=1\linewidth]{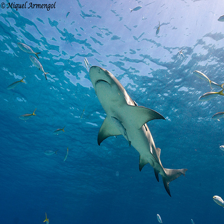}
        \caption{}
        \label{fig:moti_a}
    \end{subfigure}
    \begin{subfigure}{0.11\textwidth}
        \centering
        \includegraphics[width=1\linewidth]{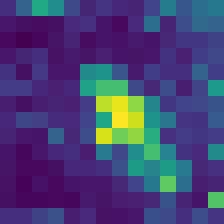}
        \caption{}
        \label{fig:moti_b}
    \end{subfigure}
    \begin{subfigure}{0.11\textwidth}
        \centering
        \includegraphics[width=1\linewidth]{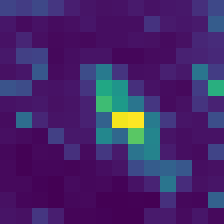}
        \caption{}
        \label{fig:moti_c}
    \end{subfigure}
    \begin{subfigure}{0.11\textwidth}
        \centering
        \includegraphics[width=1\linewidth]{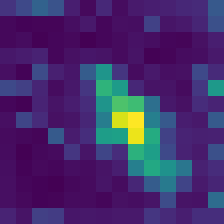}
        \caption{}
        \label{fig:moti_d}
    \end{subfigure}
    \caption{\textbf{Visulization of FAMT.} (a) is the original image. (b)-(d) are visualizations for the self-attention of the \texttt{[CLS]} token on the heads of the last layer following DINO~\cite{dino}, which denotes the results of different masking and throwing schemes based on MAE. (b) by random masking strategy, (c) by frequency \& attention-driven masking strategy, and (d) by frequency \& attention-driven masking and throwing strategy.}
    \label{motivation}
\end{figure}

The strategy of masking is highly crucial for MIM, and researchers have attempted to investigate various methods for masking to achieve better performance. To mask an image, different techniques are explored, such as random, block-wise, and grid-wise masking, as demonstrated in works like MAE~\cite{mae}, MaskFeat~\cite{ MaskFeat} and SimMIM~\cite{SimMIM}, on the other hand, applies a variety of masked patch sizes to determine the most effective size for masking. In addition to the masking method, the impact of the masking ratio has also been studied in most of the approaches mentioned above. Impressively, MIM techniques have shown exceptional performance with high masking ratios, such as 75\% for MAE and 60\% for SimMIM.

While random masking may seem a viable approach for MIM, significant issues exist that need to be addressed. Each image block has the same probability of being masked since random masking treats them equally. It is obvious that random masking tends to disperse attention throughout the entire image rather than focusing on the object of interest. However, not all the image blocks require elaborate representation learning, such as the background. Thus, random masking could make the model waste attention on irrelevant background elements and produce weaker representations, as illustrated in Fig.~\ref{fig:moti_b}.
Furthermore, random masking without focused areas always requires substantial computing resources for pretraining in MIM, which leads to massive training costs.


\revisecolor{To tackle the aforementioned issues, it is natural to utilize the attention map as guidance for masking during the pre-training phase, which achieves effective semantic information extraction. However, the attention map alone cannot sufficiently help models focus on objects of interest, where some salient features might be either overly focused or ignored. 
To alleviate the gap between attention maps and objects of interest, we further incorporate frequency domain information.
Several studies have highlighted that the self-attention mechanism in Vision Transformers (ViTs) functions similarly to a low-pass filter, contrasting with CNN models \cite{parkvision,wanganti}.
Frequency domain information can effectively guide the network's feature extraction.}

\revisecolor{In this paper, we introduce the \textbf{Frequency \& Attention-driven Masking and Throwing Strategy} (FAMT) to tackle both of the issues mentioned above. At first, we 
obtain an attention map as a constraint for masking, which we refer to as semantic information extraction, in a completely unsupervised manner. We then introduce frequency domain information to supplement the semantic information provided by the attention map. To be specific, we low-pass filter the image token and then assign a weight to each token corresponding to the image block according to the component. The image patch that has more low-pass components will get a higher weight. After that, we combine the attention values to the final weights of each image patch. The greater the weight of the image patch is, the more likely the patch will be masked. 
Such operation guarantees that the masked parts are highly informative regions related to the object, making the whole reconstruction prediction task more difficult. To further reduce computational costs, we discard regions with medium weights according to our designed strategy. As shown in Fig.~\ref{motivation}, our method can also significantly reduce the distribution of attention on the background, and the throwing operation can further smooth the attention distribution. }


The proposed FAMT module can be effortlessly integrated into MIM frameworks like MAE and SimMIM. By using the self-attention mechanism, semantic information can be acquired. Our method speeds up the pre-training process significantly due to the proposed throwing strategy. Additionally, the performance of our approach outperforms the original MAE. Specifically, our strategy boosts the linear probing accuracy of MAE by $1.3\% \sim 3.9\%$ on various datasets, including CIFAR-10, CIFAR-100, Tiny ImageNet, and ImageNet-1K. Moreover, the fine-tuning accuracy of MAE is also enhanced. On top of that, our approach achieves exceptional outcomes on prevalent downstream detection and segmentation tasks such as COCO~\cite{coco} and LVIS~\cite{lvis}.
\begin{figure*}[ht]
    \centering
    \includegraphics[width=\textwidth]{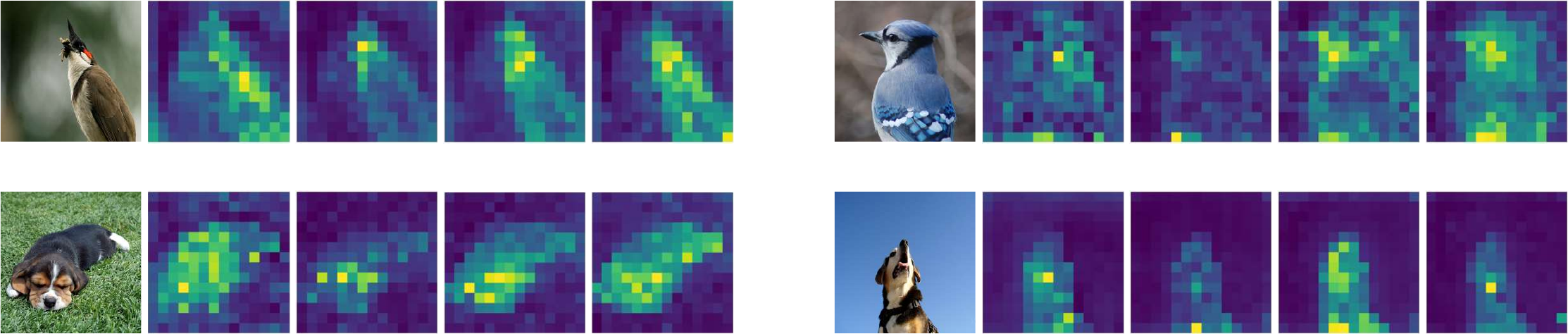}
    \caption{Visualization of attention. For each subfigure, reading from left to right and top to bottom, there are the following images: the original image and the attention map from the last layer of the MAE encoder at different training stages (\textit{40th, 60th, 80th, 100th}).}
    \label{fig:attention_visual}
\end{figure*}
The contribution of this paper can be summarized as follows:
\revisecolor{\begin{itemize}
    \item We propose an unsupervised approach that leverages the self-attention mechanism to compute attention maps during training, allowing for the extraction of token weight information. This method is applicable to any MIM method.
    \item We incorporate frequency domain information to enrich the token importance metrics. Building on the overall token importance, we design masking and discarding strategies that not only enhance performance but also reduce computational overhead.
    \item Our approach, FAMT, can be seamlessly integrated as a plug-and-play training method with any Masked Image Modeling (MIM) technique. It consistently achieves significant performance improvements across various datasets, demonstrating its strong generalizability while also reducing computational overhead.
\end{itemize}}
\revisecolor{This paper is a follow-on work of our previous conference paper\cite{liu2023good}. In terms of the methodology, we transform each image token to the frequency domain by Fast Fourier Transformation (FFT) in the rounds where the sampling weights are updated periodically, and get the weight of the low-frequency component in the Direct-Current (DC) component $\gamma$ \cite{antioversmooth} for different image patches after passing through a low-pass filter. The $\gamma$ and attention weights are summed to obtain the final sampling weights of the image token, which are used to guide the masking and throwing operations. Compared to the conference version, FAMT has incorporated the image semantics of Vision Transformers (ViT) from a frequency domain perspective, thereby further enhancing the quality of token sampling during self-supervised pre-training. This approach has demonstrated superior performance across multiple tasks, with experimental results available in Section \ref{sec:exp}.} The details are described in Section \ref{sec:method}. In addition, we use a different backbone from the conference version for the experiments and extend the training period, which demonstrates the universality and scalability of the method to a certain extent. In addition to the dataset used in the conference, we also validate the method on the remote sensing segmentation dataset iSAID\cite{1isaid, 2isaid}. Visualizations of the segmentation results further demonstrate that the method can improve the accuracy of the segmentation task. The details are described in Section \ref{sec:exp}. In addition, we conducted ablations on the newly proposed frequency domain-based method for filtering image blocks to demonstrate the effect of each module on the performance, and the specific analysis is given in Section \ref{sec:abl}.

\section{Related Work}
In this section, we present significant previous works that are pertinent to our topic. Specifically, we discuss the approaches of self-supervised learning, masking strategy, and frequency domain information.

\subsection{Self-supervised Learning}

\revisecolor{\subsubsection{Contrastive learning} Contrastive learning has been the dominant self-supervised learning paradigm for a considerable time. Its fundamental objective is to bring positives together while pushing negatives apart~\cite{he2019moco,chen2020mocov2,chen2021mocov3,chen2020simclr,grill2020byol,dino,OCTcon,Arbitrarycon,videocon,videocon2}. However, the method of sampling data views remains a challenging issue, and several works have attempted to address it~\cite{DLSL,pretext1}. One approach is to sample based on the importance of image views, which guides the generation of positives and negatives~\cite{contrastivecrop,CAST}. In our method, we leverage the self-attention mechanism along with frequency domain information to extract the importance of tokens. Contrastive learning has also been used in downstream tasks like Action Recognition {\em etc.}\cite{action}}

\noindent
\subsubsection{Masked language modeling (MLM)} Transformers have achieved significant success in natural language processing (NLP), particularly in pre-training, with methods such as BERT~\cite{bert} and GPT\cite{gpt3}. These models use MLM, where they predict concealed content based on only a limited portion of the input sequence. Pre-training these models on extensive data has demonstrated their scalability across a range of downstream tasks, indicating the strong generalization ability of MLM.

\begin{figure*}[t]

    \centering
    \includegraphics[width=0.8\textwidth]{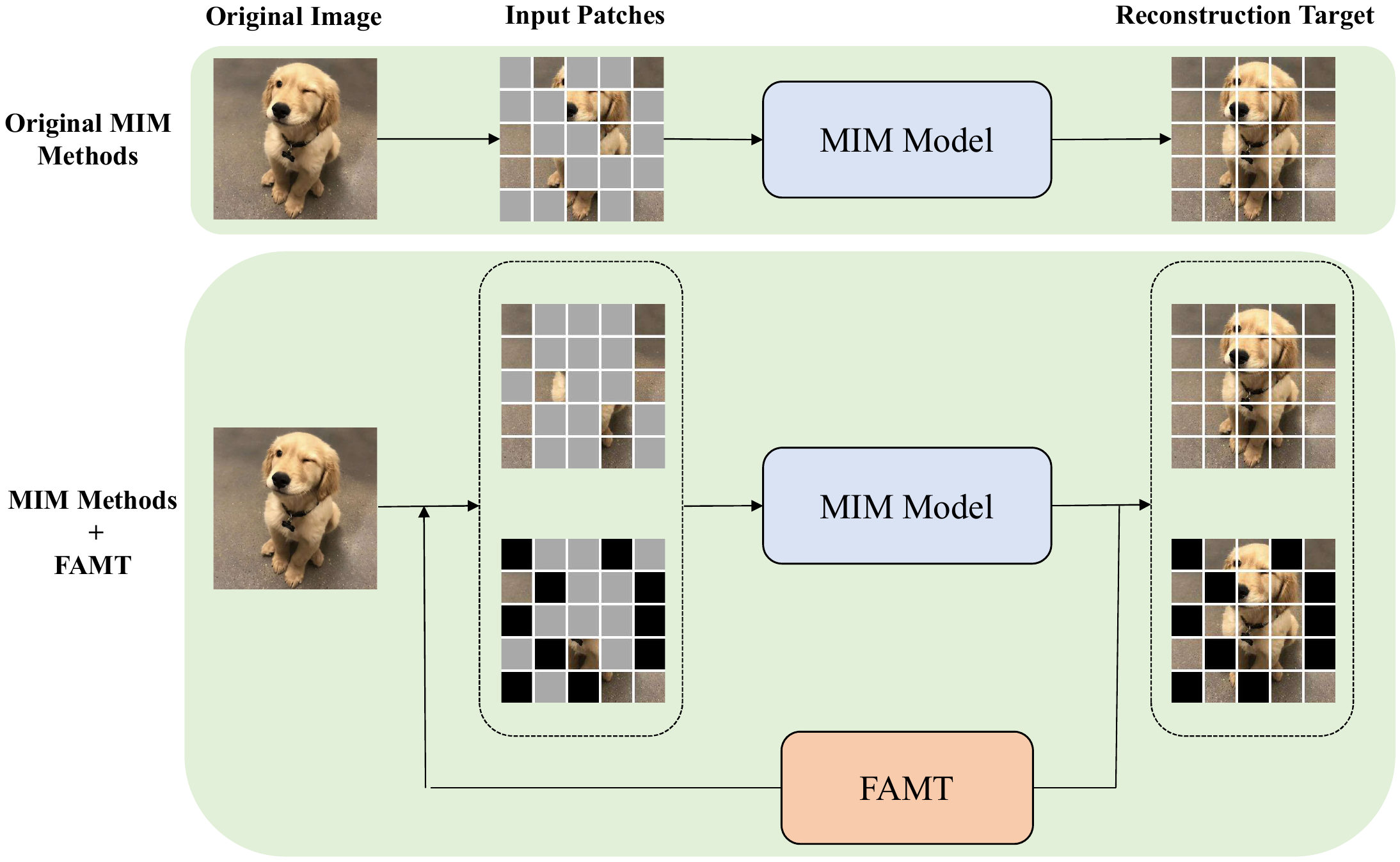}
    \caption{Overview of common MIM methods and FAMT. The top of the figure denotes the simplified common MIM methods and the bottom is the simplified overview of our FAMT. The \colorbox{lightgray}{gray} patches are masked patches. The \colorbox{black}{\textcolor{white}{black}} patches denote thrown tokens that are not input into the model, meaning that thrown tokens do not cost computational resources. Compared to original methods, FAMT leverages the frequency information and attention to mask and throw intentionally. }
    \label{fig:method_visual}
\end{figure*}

\noindent
\subsubsection{Masked image modeling (MIM)}  Recently, there has been a surge of interest in MIM~\cite{igpt,doersch2015context,henaff2019CPC,pathak2016context,trinh2019selfie,videobg}. Context encoders~\cite{pathak2016context} were among the earliest works in this direction, which predicted missing pixels in specific regions. With the increasing popularity of transformers~\cite{attention,swin,swinv2,vit,cait,deit}, MIM has regained attention. iGPT~\cite{igpt} and ViT~\cite{vit} propose innovative strategies for utilizing transformers to process images. Distillation-based MIM has also emerged~\cite{ibot}. BEiT~\cite{beit} uses a trained dVAE network to construct a challenging task that predicts the visual tokens of masked image patches. Similarly, MAE~\cite{mae} employs an autoencoder for MIM, which learns representations through the encoder and reconstructs original pixels of masked image patches through the decoder. Unlike MAE, SimMIM~\cite{SimMIM} and MaskFeat~\cite{MaskFeat} utilize a linear head in place of a transformer decoder. MaskFeat substitutes original pixels with HOG~\cite{hog} features as the target for reconstruction.

\subsection{Masking Strategy}

\subsubsection{Random masking} MIM heavily relies on predicting masked image patches to learn representations, highlighting the critical role of masking strategy in MIM. BEiT utilizes a block-wise random masking strategy that may mask a block of patches instead of individual patches. Block-wise masking has also been employed in~\cite{MaskFeat}. On the other hand, MAE randomly masks a large number of patches, and the size of the masked patches is the same as the input patch size of ViT ($16 \times 16$). Furthermore, SimMIM investigates the impact of various masked patch sizes and ultimately selects a larger size ($32 \times 32$).

\begin{figure}[t]
    \centering
    \includegraphics[width=0.45\textwidth]{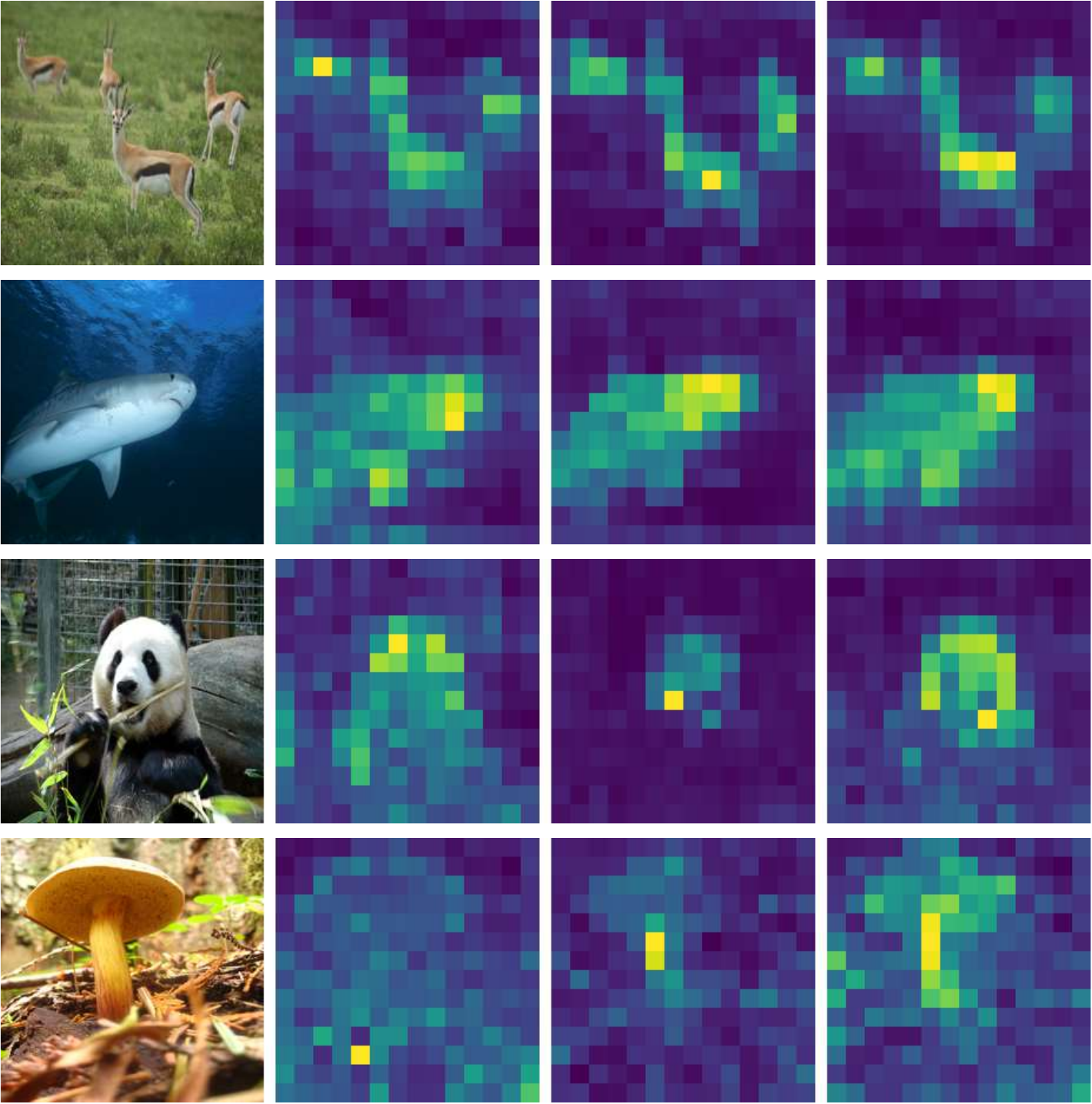}
    \caption{Visualization of the attention map of the last layer in the encoder after 400 epochs pre-training. From left to right, there is the original image, the attention map from the last layer of
    the MAE encoder using random masking, attention-driven masking, and FAMT, respectively.}
    \label{fig:result_visual}

\end{figure}

\noindent
\subsubsection{Selective masking} Instead of relying solely on random masking strategies, selective masking schemes have recently been explored in several works. MST~\cite{MST}, for example, advocates for masking low-attended patches, achieving good performance without additional cost. AttMask~\cite{whattohide} goes further by investigating the results of masking various highly-attended patches, showing the effectiveness of such an approach. However, these methods are only applicable to a specific distillation-based model. In comparison, our FAMT is a plug-and-play module that can be readily incorporated into popular MIM methods, such as MAE, SimMIM, and MaskFeat. SemMAE~\cite{semmae} has also introduced a semantic-guided masking strategy, but their approach requires an additional pre-trained model to extract features and uses these features in a complex way, resulting in increased computational resource usage. In contrast, our FAMT is a simple and framework-agnostic module that fully unsupervisedly obtains semantic information without any additional design.

\begin{figure}[ht]
    \centering
    \includegraphics[width=0.45\textwidth]{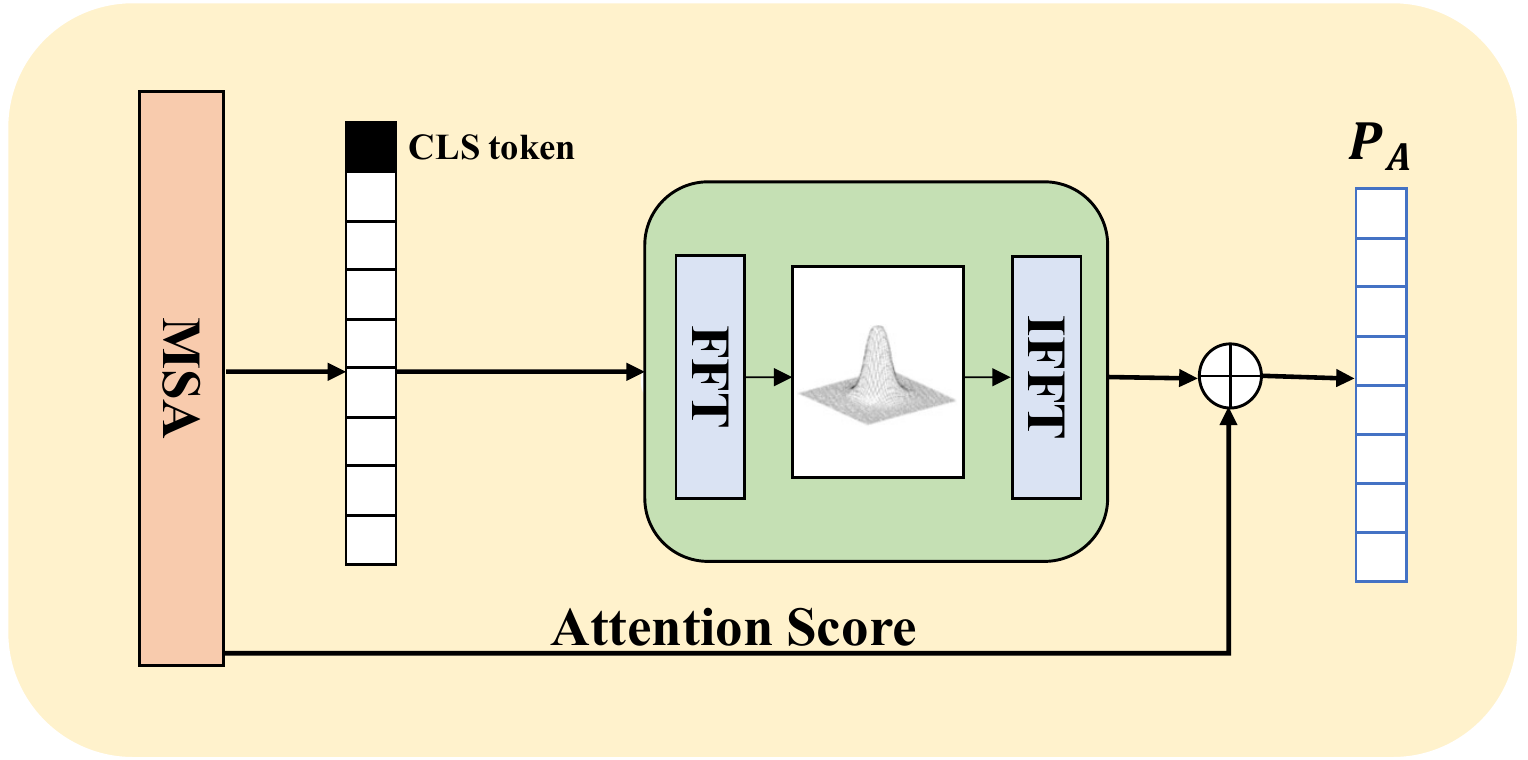}
    \caption{The pipeline of FAMT for updating $P_A$. The filter is a Gaussian low-pass filter.}
    \label{fig:FAMTforpa}

\end{figure}

\begin{figure}[ht]
    \centering
    \includegraphics[width=0.45\textwidth]{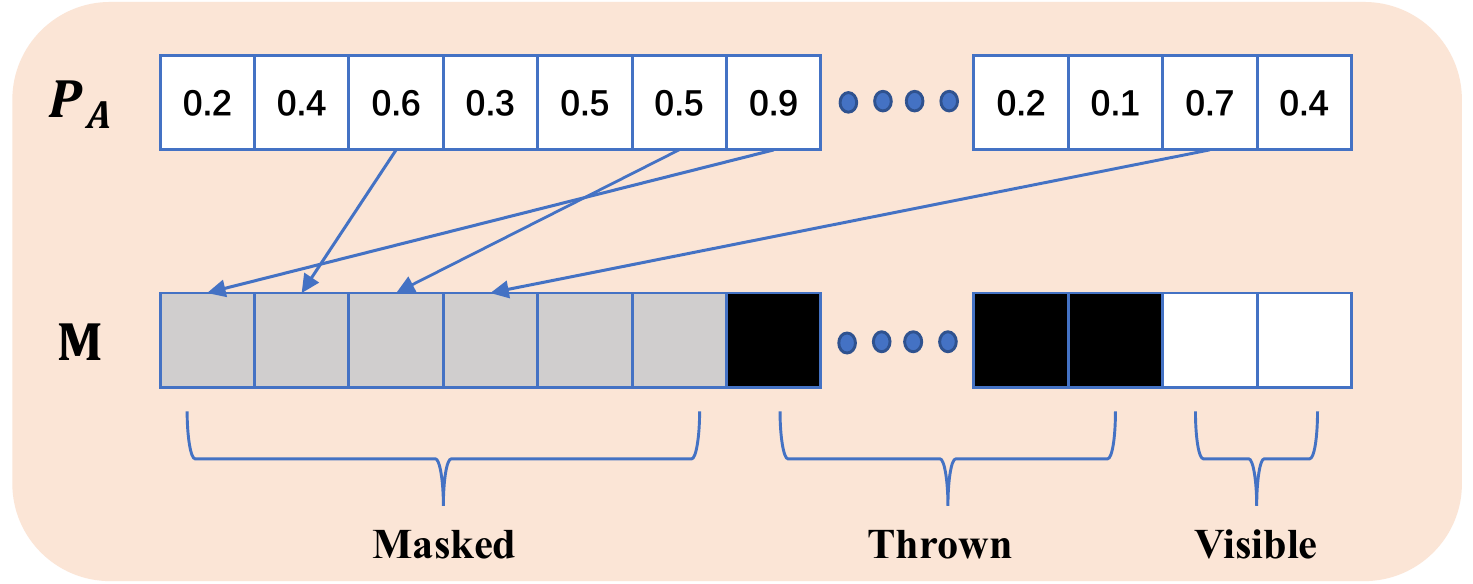}
    \caption{The pipeline of sampling from $P_A$. M denotes the final index set of patches. The gray areas are the mask regions, and the black ones are the throw regions.}
    \label{fig:sample_visual}

\end{figure}

In addition to implementing a masking strategy, our proposal for the FAMT includes a throwing strategy. By utilizing this approach, we are able to improve overall performance while also significantly reducing computing costs.

\subsection{Frequency domain information} 

\revisecolor{Recent studies, as referenced\cite{wang2020high, park2022vision, rao2021global, antioversmooth, mproving} have demonstrated that Vision Transformers exhibit a contrasting behavior to Convolutional Neural Networks within the frequency domain. Considering the redundancy of the image information, we also try to use the frequency domain information to filter the image blocks.
\cite{antioversmooth} generalizes the self-attention mechanism as a low-pass filter using the Fourier spectrum domain. \cite{mproving} verifies a hypothesis that ViT models perform worse than CNN models in utilizing the high-frequency components of the images by frequency analysis. Apart from these, \cite{mproving} also indicates that ViT models are more prone to capture the low-frequency parts of the images and thus get better performance than most CNN models. In addition, compared to contrastive learning, MIM artificially leverages high-frequency information \cite{park2023self}.}

\section{Method}
\label{sec:method}
In this section, we will introduce \textit{frequency \& attention-driven masking and throwing strategy} (FAMT) for MIM. Our paper begins by laying out foundational concepts of vision transformer and MIM. We then proceed to provide a detailed account of FAMT, which encompasses the collection of semantic data, masking approach, and throwing policy.

\subsection{Preliminary}

\subsubsection{Revisiting Vision Transformer}
\label{sec: ViT}


In this section, we introduce ViT~\cite{vit}, a widely used architecture that treats images as sequences of tokens. To be specific, an input image $x \in \mathbb{R}^{H\times W \times C}$ is reshaped into a 1D sequence of token embeddings $x_p \in \mathbb{R}^{N \times(P^2 \cdot C)}$, where $N = HW/ P^2$ is the number of patches and $(P, P)$ represents the size of each image patch. Next, the patches are projected to a $D$-dimensional space through a linear projection $E \in \mathbb{R}^{(P^2 \cdot C) \times D}$, and a special \texttt{[CLS]} token $x_{cls}$ is added to the sequence. Finally, a position embedding $E_{pos} \in \mathbb{R}^{(N+1) \times D}$ is included to create the tokenized image input:
\begin{equation}
    z = [x_{cls}; x_p^1E; x_p^2E;...; x_p^NE] + E_{pos},
\label{eq: image token1}
\end{equation}where $x_p^m$ represents the $m$-th row of $x_p$, and $z \in \mathbb{R}^{(N+1) \times D}$ can be used as the input for ViT.

\subsubsection{Revisiting masked image modeling}

The patch-level image processing in ViT allows for the independent handling of each image patch, enabling patch masking. The commonly used MIM methods, such as MAE, SimMIM, and MaskFeat, are referred to as original MIM methods, as depicted in Fig.~\ref{fig:method_visual}. \revisecolor{These MIM methods update the model weights by predicting the masked parts of the image, using $l_1$ or $l_2$ losses as follows:
\begin{equation}
L_p = \parallel Y_M - X_M \parallel_{p},
\end{equation}
where $Y$ and $X$ represent the predicted values and the features to predict (such as RGB pixels or other features, {\em e.g.}, HOG features), respectively, and $M$ denotes the corresponding mask. Note that the loss is only computed on masked patches.}

\subsection{Frequency \& Attention-Driven Masking and Throwing}

\subsubsection{Semantic information extraction}
\label{sec: semantic inform}
\revisecolor{After the image is tokenized as $z \in \mathbb{R}^{(N+1) \times d}$ (as shown in Eq. (\ref{eq: image token1})), the tokens are fed into a Transformer block. The Transformer block utilizes a \textit{multi-head self-attention} (MSA) layer to divide $z$ into $h$ heads, each containing query $q_i$, key $k_i$, and value $v_i$ for $i = 1, 2,...,N_h$. Here, $q_i,k_i,v_i \in \mathbb{R}^{(N+1) \times d}$. Softmax is then applied to obtain the MSA as follows:
\begin{equation}
A = softmax(q_ik_i/ \sqrt{d/h}),
\end{equation}
where $A$ represents the $(N+1) \times (N+1)$ attention matrix. After obtaining $A$, the first row (excluding the first element) is averaged over $h$ heads to get $a_w$ as follows:
\begin{equation}
a_w = \frac{1}{h} \sum\limits_{i=1}^{h} a^1,
\end{equation}
where $a^1 \in \mathbb{R}^N$ is the first row of $A$, i.e., the [CLS] attention distribution,
without the first element.}
Then, $a_w$, which is referred to as masking weights, is reshaped to $(H/P) \times (W/P)$ and mapped to the original image size using interpolation. This process helps the model to capture semantic information roughly, even at the early stage of pre-training.
During pre-training, $a_w$ is updated every 40 epochs. Precise object location is unnecessary since rough location provides enough semantic information to guide masking and throwing. Moreover, this forward step incurs only a trivial computing cost and can be overlooked (accounting for about $1\%$ of the entire pre-training time).

\begin{algorithm}[bt]
    \caption{Algorithm of FAMT for MIM}
    \label{alg:algorithm}
    \textbf{Input}: Image token \textit{Z}, Masking ratio \textit{r}, Throwing ratio \textit{t}, Height of Image \textit{H}, Width of Image \textit{W}, Patch size \textit{P}\\
    \begin{algorithmic} 
    \State $N = (H \times W) / P^2$ \Comment{The number of patches}
    \State $C_{m} = N \times r$ \Comment{The count of masked tokens}
    \State $C_t = N \times t$ \Comment{The count of thrown tokens}
    \State $a_w = Forward(Z)$ \Comment{Attention map of last layer}
    \State $\mathcal{Z} = \mathcal{F}(Z)$ \Comment{FFT}
    \State $\gamma_{j} = \frac{\parallel \mathcal{LC}[\mathcal{Z}_{j,:}] \parallel_2}{\parallel \mathcal{DC}[\mathcal{Z}_{1:,:}] \parallel_2}$ \Comment{Frequency domain weights}
    \State $P_{A_{_{i}}} = \frac{\gamma_i \odot a_{w_{i}}}{\sum_{j}^{N} \gamma_j \odot a_{w_{j}}}$ \Comment{Sampling weights}
    \For{$k = 1 \to N$}
        \State $M[k] = \mathbb{I}(U;P_A)$ \Comment{Sampling by Eq. (\ref{eq:sampling})}
    \EndFor
    \State $mask\_idx = M[:C_m]$  \Comment{Masking tokens}
    \State $throw\_idx = M[C_m:C_m+C_t]$ \Comment{Throwing tokens}
    \end{algorithmic}
    
    \textbf{Output}: $mask\_idx,throw\_idx$
    \end{algorithm}

\subsubsection{Frequency \& Attention based selection}
\label{sec:frequency}

\revisecolor{
Apart from the self-attention in the Transformer block, we also propose to incorporate frequency information into the computing of the sampling weight for masking and throwing. First, we use $Z$ to denote the output of the Transformer block. Taking inspiration from \cite{antioversmooth, vtclfc}, we assess the low-frequency part of tokens $Z\in \mathbb{R}^{(N+1) \times d}$ by utilizing FFT on each channel of tokens except for the \revisecolor{[CLS]} token to transform them into the frequency domain, which is represented as $\mathcal{Z}_{1:,:} = \mathcal{F}({Z_{1:,:})}$. Following \cite{vtclfc}, we apply a low-pass filter $\mathcal{G}$ with cutoff factor $\sigma$ to obtain the proportion of the low-frequency component in the total DC component. Then IFFT is used to recover tokens from frequency domain to spatial domain. Thus we can get a score $\gamma$ in a range of $[0,1]$, denoted as
\begin{equation}
\gamma_{j} = \frac{\parallel \mathcal{LC}[\mathcal{Z}_{j,:}] \parallel_2}{\parallel \mathcal{DC}[\mathcal{Z}_{1:,:}] \parallel_2}  = \frac{\parallel \mathcal{F}^{-1}(\mathcal{G}(\sigma) \odot \mathcal{Z}_{j,:}) \parallel_2}
{\parallel \mathcal{F}^{-1}(\mathcal{Z}_{1:,:}) \parallel_2},
\end{equation}
where $j$ is the token index, which does not include the [CLS] token. $\odot$ denotes the hadamard product. LC means that the low-frequency component of the total frequency component DC.
Furthermore, we update the value in $a_w$ leveraging $\gamma$.
Specifically, we balance the original self-attention mechanism’s attention weights $a_w$ with frequency domain weight information $\gamma$. This approach allows our weighting mechanism to incorporate both high-level semantic information and low-level energy details. Finally, we get the weight $P_A$ for masking and throwing: 
\begin{equation}
    P_{A_{_{i}}} = \frac{\gamma_i \odot a_{w_{i}}}{\sum_{j}^{N} \gamma_j \odot a_{w_{j}}}.
\end{equation}
}

\subsubsection{Masking and Throwing}
Fig.~\ref{fig:method_visual} illustrates that original MIM methods employ random masking, which gives equal chances to all image patches to be masked. This operation can cause the model to disperse its attention across the entire image, which damages representation learning, as demonstrated in Fig.~\ref{fig:result_visual}.
\revisecolor{
To address this issue, we propose to mask the top important tokens only and leave certain parts of the object visible (Fig.~\ref{fig:method_visual} \textit{masking only}). This generates a more challenging reconstruction task, as the model has to focus on the less-attended and high-frequency regions.
However, direct sampling can lead to a bias towards ranking highly-attended \& low-frequency patches at the top and low-attended \& high-frequency patches at the bottom. }

\revisecolor{To address these challenges, we propose a method that intentionally masks and discards image patches. Specifically, we first use random masking to gather semantic information from the entire image during the early stages of pre-training. After a few epochs, we use the method described in this section to obtain an attention map, which is mentioned in Subsection III-B2. Each element in \( P_A \) represents the weight of the corresponding pixel in the image.
We employ the Inverse Transform Sampling strategy \cite{wiki:inverse_transform_sampling} to ascertain the 
\( N \) patch indices to get a mask \(M\). This is mathematically delineated as follows:
\begin{equation}
F(i) = \sum_{k=1}^i P_{A_k}, \quad i \in {1, \dots, N},
\end{equation}
\begin{equation}
\label{eq:sampling}
I = \mathbb{I}(U; F(i)), \quad U \sim \text{Uniform}(0, 1).
\end{equation}
Here, the cumulative distribution function \( F(i) \) is derived from \( P_A \). The inverse transform sampling function \( \mathbb{I} \) utilizes a uniformly distributed random variable \( U \) to sample indices from \( F(i) \). Note that we ensure the sampling is non-repetitive. This ensures a probabilistic congruence with the importance of each token, thereby enhancing the sampling diversity and alignment with the statistical properties of \( P_A \).
}

 \revisecolor{ We do not directly mask highly-attended \& low-frequency areas but increase the likelihood of these patches being masked. This ensures that even highly-attended \& low-frequency regions have a small probability of being visible. Fig.~\ref{fig:result_visual} shows that the attention of the model is more focused on minor areas that contain salient features, but it may also overlook several significant parts of the object ({\em e.g.}, the body of the panda, the head of the mushroom, {\em etc.}). As a result, the model's ability to learn representations may be compromised.}

\revisecolor{Additionaly, random masking typically requires significant time due to the large size of the backbone and huge amount of data.
To reduce computation overload, we introduce the throwing strategy, which leverages the \( P_A \) to guide the throwing of tokens. Since we can estimate the location of the semantic object from the \( P_A \), we can safely discard parts of the original sample that are not informative for training. Specifically, we randomly remove a certain number of tokens in the middle of $M$, according to the throwing ratio $t$ and masking ratio $r$, with the top tokens being masked and the bottom tokens being visible. The resulting tokens are denoted as $z_I$ which contains both masked and visible parts. Such a process is visualized in Fig.~\ref{fig:sample_visual}.
As shown in Fig.~\ref{fig:result_visual}, the FAMT strategy improves the model's ability to focus on salient regions while decreasing its attention on the background. Furthermore, the FAMT approach promotes a smoother transition between salient and common features, such as the head and body of the object.}

The FAMT strategy can be easily integrated into existing MIM methods. We explore the impact of throwing different areas on the performance of the model in Section~\ref{sec:abl}. Additionally, the throwing strategy allows us to significantly reduce the size of the input data, resulting in faster pre-training. Algorithm~\ref{alg:algorithm} provides an overview of the FAMT pipeline.

\begin{table*}[ht]
        \centering
        \small
        \renewcommand\arraystretch{1.2}
        \setlength\tabcolsep{3pt}
            \begin{tabular}{lcccccccccccc}
                \toprule
                \multicolumn{1}{c}{\multirow{2}{*}{Method}}  & \multicolumn{2}{c}{CIFAR-10}                    & \multicolumn{2}{c}{CIFAR-100}                   & \multicolumn{2}{c}{Tiny ImageNet}                                   & \multicolumn{2}{c}{ImageNet} \\ \cmidrule(r){2-3}\cmidrule(r){4-5}\cmidrule(r){6-7}\cmidrule(r){8-9}\cmidrule(r){10-11}
                \multicolumn{1}{c}{}                            & Linear                 & Fine-tuning            & Linear                 & Fine-tuning            & Linear                 & Fine-tuning            & Linear                 & Fine-tuning              \\ \midrule
                MAE+Random M     & 76.2   & \underline{98.2}    & 51.9    & 87.4   & 46.8    & \underline{77.7}  & \underline{47.4}  & \bf80.6            \\
                MAE+FAM (ours)  & \bf79.9 & \bf98.2   & \underline{55.7} & \underline{87.5}  & \bf49.4 & \bf78.0  & \bf48.8   & \bf80.6            \\
                MAE+FAMT (ours)   & \underline{79.5}   & 98.1 & \bf55.8  & \bf87.6 & \underline{48.4}  & \underline{77.7}  & 47.0  & \underline{80.4}     \\ 
                \bottomrule
                \end{tabular}
        \caption{Top-1 accuracy on CIFAR-10/100, Tiny ImageNet, and ImageNet. M denotes Masking. FAM denotes frequency \& attention-driven masking. Random Masking is the default masking strategy for MAE.}
        \label{tab:classification800s}
        \end{table*}

 \begin{table}[]
\centering
\small
\renewcommand\arraystretch{1.2}
\begin{tabular}{@{}lccc@{}}
\toprule
\multicolumn{1}{c}{MAE} & CIFAR-10 & CIFAR-100 & ImageNet \\ \midrule
+Random masking         & 73.5     & 46.6      & \underline{47.0}     \\
+FAM                    & \bf77.5  & \underline{51.6}      & \bf48.6     \\
+FAMT                   & \underline{77.1}     & \bf52.2   & 46.8     \\ \bottomrule
\end{tabular}
\caption{Top-1 accuracy on 70\% CIFAR-10/100 and ImageNet. All models are pretrained on ImageNet-1K.}\label{tab:linearclassification070}               
\end{table}

\begin{table}[]
\centering
\small
\renewcommand\arraystretch{1.2}
\begin{tabular}{@{}lccc@{}}
\toprule
\multicolumn{1}{c}{Method} & aAcc & mIOU & mAcc \\ \midrule
Random Init                & 65.8 & 16.2 & 22.0 \\
MAE$^\dag$               & 79.2 & 38.1 & 49.0 \\
MAE+FAM                    & \bf79.9 & \underline{39.7} & \bf50.5 \\ 
MAE+FAMT                   & \underline{79.7} & \bf39.8 & \underline{50.4} \\ \bottomrule
\end{tabular}
\caption{The results on ADE20K. The models are pretrained on ImageNet-1K. $^\dag$: default MAE with random masking.}
\label{tab:segade}  
\end{table}

\begin{table}[t]
\centering
\small
\renewcommand\arraystretch{1.2}
\begin{tabular}{@{}lccc@{}}
\toprule
\multicolumn{1}{c}{Method} & aAcc    & mIOU   & mAcc     \\ \midrule
MAE$^\dag$                    & \underline{98.7}    & \underline{58.9}   & \underline{66.9}    \\
MAE+FAM              & \bf98.8 & \bf60.0   & \bf67.7     \\
MAE+FAMT             & \underline{98.7}    & 57.2   & 64.8   \\ \bottomrule
\end{tabular}
\caption{The results on iSAID. $^\dag$: default MAE with random masking.}
\label{tab:segisaid}  
\end{table}

\section{Experiments}
\label{sec:exp}

\subsection{Setup}

\subsubsection{Datasets and Baseline Approaches}

Our method is evaluated with popular MIM technique (MAE) through linear probing and fine-tuning on ImageNet-1K validation set. We further validate the transferability of our approach on other downstream tasks including classification accuracy on datasets such as \textbf{CIFAR-10/100}\cite{cifar}, \textbf{Tiny ImageNet}, and \textbf{ImageNet-1K}\cite{imagenet} by employing linear probing and fine-tuning. Additionally, we fine-tune our method on     \textbf{COCO}\cite{coco} and \textbf{LVIS}\cite{lvis} datasets for object detection and segmentation. Additionally, we also validate the segmentation performance of our method on \textbf{ADE20K}\cite{1ade20k,2ade20k} and \textbf{iSAID}. Furthermore, we provide ablation studies on the ratio and the part of masking and throwing.

\subsubsection{Implementation Details}
Patch-level masking is a widely used technique in MIM methods for providing self-supervised signals, making it a suitable plug-and-play module for MIM methods. FAMT is designed to be independent of other training components, such as losses, optimizers, and learning rate schedules. To ensure a fair comparison with the original MIM method, we maintain the same training settings for both methods. This comparison aims to evaluate the performance improvement achieved by FAMT.

We conducted an 800-epoch pretraining of MAE on ImageNet-1K using the original pretraining settings. Our encoder backbone was ViT-S/16, and our decoder used ViT-S/16 with 8 blocks and an embedding dimensionality of 128. In all experiments, we utilized absolute position embedding. To evaluate classification with MAE, we implemented a revised linear head with batch normalization for linear probing. The same settings as the original MAE were used for fine-tuning, with the \texttt{[CLS]} token used for both linear probing and fine-tuning.

We have selected a throwing ratio of $t=0.4$ for our FAMT approach applied to MAE. To maintain a consistent ratio between masked and visible tokens, we set the masking ratio to $r=0.45$. This ensures that the different ratios between masked and visible tokens do not affect the results. To update the masking weights $a_w$, we set an interval of 80 epochs, which is equivalent to $10\%$ of the entire pre-training process for MAE. The additional evaluation step has a negligible cost, and our FAMT method accelerates the training process with the help of the throwing strategy. All of our experiments were conducted on an 8-GPU server.

\subsection{Classification }
In this section, we present the results of pretraining MAE on ImageNet-1K using diverse masking and throwing strategies for 800 epochs. We evaluate the performance of the pretrained model on different datasets using both linear probing and fine-tuning techniques. The datasets we consider include CIFAR-10/100, Tiny ImageNet, ImageNet-1K, and STL-10.

\subsubsection{Linear probing}

Tab.~\ref{tab:classification800s} demonstrates the linear probing performance of MAE, and the results indicate that our FAMT significantly enhances the Top-1 accuracy by $1.6\% \sim 3.9\%$ on CIFAR-10/100, Tiny ImageNet. This highlights the remarkable improvement of our FAMT on the linear separability of learned representations. The transferability of FAMT is obviously higher than that of the original MAE. As for ImageNet, FAM gets better performance than MAE, but FAMT has a little performance loss. Intuitively, we think that such an operation causes the model to lose classification power on the specified pre-trained dataset, but does not affect the migration ability of the model. The capacity to perform well on downstream tasks is the primary objective of self-supervised pre-training, making it a valuable capability. It is noteworthy that FAMT achieves faster pre-training by using only $60\%$ of the image.

Additionally, we linear probe on 70\% CIFAR-10/100 and ImageNet. The results are shown in Tab.~\ref{tab:linearclassification070}. Compared to the original method, FAM achieves obvious improvements on all datasets. FAMT gets better performance on CIFAR-10/100, and notably gains the best result on CIFAR-100. One important point is that FAM can get better results than original MAE with only 70\% of dataset. More specifically, both FAM and FAMT got better performance than original MAE when the data volume drops. This is partly an indication that FAM and FAMT use the data more effectively.

\subsubsection{Fine-tuning}

Fine-tuning accuracy could reflect the strength of the learned non-linear features, which is important for downstream tasks.
The results of fine-tuning are shown in Tab.~\ref{tab:classification800s}. Our FAMT gets competitive performance with the original method.   Besides, we experimentally find the choice of hypermeters for our FAMT is more extensive than original methods when fine-tuning, which reduces the time for searching appropriate \textit{lr}.

\begin{table*}[ht]
    \centering
    \small
    \renewcommand\arraystretch{1.2}
    \setlength\tabcolsep{3pt}
    \begin{tabular}{lccccccllllll}
        \toprule
        \multicolumn{1}{c}{\multirow{2}{*}{Method}}             & \multicolumn{6}{c}{\textit{LVIS}}      &\multicolumn{6}{c}{\textit{COCO}}                                                                                                                             \\\cmidrule(r){2-7} \cmidrule(r){8-13}
        \multicolumn{1}{c}{}                         & AP$^{bbox}$ & AP$_{50}^{bbox}$ & AP$_{75}^{bbox}$ & AP$^{mask}$ & AP$_{50}^{mask}$ & AP$_{75}^{mask}$  &\multicolumn{1}{c}{AP$^{bbox}$} & \multicolumn{1}{c}{AP$_{50}^{bbox}$} & \multicolumn{1}{c}{AP$_{75}^{bbox}$} & \multicolumn{1}{c}{AP$^{mask}$} & \multicolumn{1}{c}{AP$_{50}^{mask}$} & \multicolumn{1}{c}{AP$_{75}^{mask}$} \\ \midrule
    Random Init                  & 14.6   &  24.7    &  15.3   & 14.3  &  23.2    &   15.0    &28.1                       & 46.1                        & 29.8                         & \underline{26.2}                      & 43.5                        & 27.6                       \\
    MAE$^\dag$                         & 25.0   & 38.5     & 26.9     & 24.2   & 37.0     & 25.7      &40.1                       & 60.3                         & 43.7                         & \underline{36.6}                       & 57.6                         & 39.3                         \\
    MAE+AM & 26.2   & \underline{40.2}     & \underline{28.3}     & \underline{25.4}   & \underline{38.4}     & \underline{27.1}      &\underline{42.2}                       & \underline{62.6}                         & \underline{46.3}                         & \bf38.3                       & \underline{59.7}                         & \underline{41.4}                         \\
    \rowcolor[HTML]{e2e1e1}
    MAE+AMT                      & \bf26.9   & \bf41.3     & \bf28.8     & \bf26.0   & \bf39.4     & \bf27.7      &\bf42.8                       & \bf63.0                         & \bf47.1                         & \underline{36.6}                       & \bf60.1                         & \bf41.6                         \\
    
    \bottomrule
    \end{tabular}
    \caption{The results on COCO and LVIS. The models are pretrained on ImageNet-1K. AM denotes attention-driven masking. AMT denotes AM and throwing. $^\dag$: default MAE settings with random masking.}
    \label{tab:instance}
    \end{table*}

\begin{table*}[ht]
    \centering
    \small
    \renewcommand\arraystretch{1.2}

    \begin{tabular}{lcccccc}
        \toprule
        \multicolumn{1}{c}{\multirow{2}{*}{Method}}                                            & \multicolumn{2}{c}{Ratio (\%)} & \multirow{2}{*}{\makecell[c]{\textit{Attention-driven} \\  \textit{Masking}}} & \multirow{2}{*}{\textit{Throwing}} & \multirow{2}{*}{\makecell[c]{\textit{Throwing middle} \\ \textit{Tokens}}} & \multirow{2}{*}{\textit {Acc. (\%)}} \\ 
        \multicolumn{1}{c}{}                                           & Mask              & Throw              &                                                    &                                    &                                                  &                                  \\ \midrule
    MAE+Random M                                     & 75                & 0                  &                                                    &                                    &                                                  & 52.3                             \\
    MAE+Attention-driven M                              & 75                & 0                  &                    \checkmark                      &                                    &                                                  & 50.1                             \\
    MAE+Random T                          & 45                & 40                 &                                                    &     \checkmark                     &                                                  & 50.8                             \\
    MAE+AMT                              & 75                & 10                 &                    \checkmark                      &     \checkmark                     &                                                  & 51.7                             \\
    MAE+AMT                              & 45                & 40                 &                    \checkmark                      &     \checkmark                     &                                                  & \underline{52.6 }                            \\
    \rowcolor[HTML]{e2e1e1} 
    MAE+AMT                              & 45                & 40                 &                    \checkmark                      &     \checkmark                     &                 \checkmark                       & \textbf{53.3}                             \\
    \bottomrule
        \end{tabular}
        \caption{Ablation of different masking and throwing strategies used in MAE. Each model is pretrained on ImageNet-1K for 200 epochs. M denotes Masking. T denotes Throwing. \textit{Acc.} is the Top-1 linear probing accuracy on Tiny ImageNet.}
        \label{tab:ablation}
    \end{table*}

\begin{table}[ht]
\centering
\small
\renewcommand\arraystretch{1.2}
\begin{tabular}{@{}lcc@{}}
\toprule
\multicolumn{1}{c}{Method} &\textit{linear acc.} & \textit{finetune acc.} \\ \midrule
MAE                        & 28.1       & \underline{67.1}         \\
MAE+AM                     & 27.6       & \underline{67.1}         \\
MAE+AMT                    & \underline{28.4}       & 65.5         \\
MAE+FAMT                   & \bf32.9       & 65.5         \\
MAE+FAM                    & \underline{28.4}       & \bf67.2         \\ \bottomrule
\end{tabular}
\caption{Ablation of FAMT where F, AM and T denote using frequency information for mask selection, attention-driven masking, and throwing, respectively.}
\label{tab:ablation pretrain on tiny}
\end{table}

\subsection{Downstream Tasks}
In this section, results on object detection and instance segmentation tasks are shown to further investigate the transferability of our method. We experiment with models pretrained on ImageNet-1K for 800 epochs. In particular, we perform instance segmentation on ADE20K and iSAID. Following mmsegmentation codebase \cite{mmseg}, we employ the same setups with a total batch size of 16 and all experiments here use upernet~\cite{Mask-rcnn} as the detector with a backbone of ViT-S/16. To maintain a fair comparison, all hypermeters are the same in each experiment.

\subsubsection{Comparisons on ADE20K} 
\begin{figure*}[h]

    \centering
    \includegraphics[width=1\textwidth]{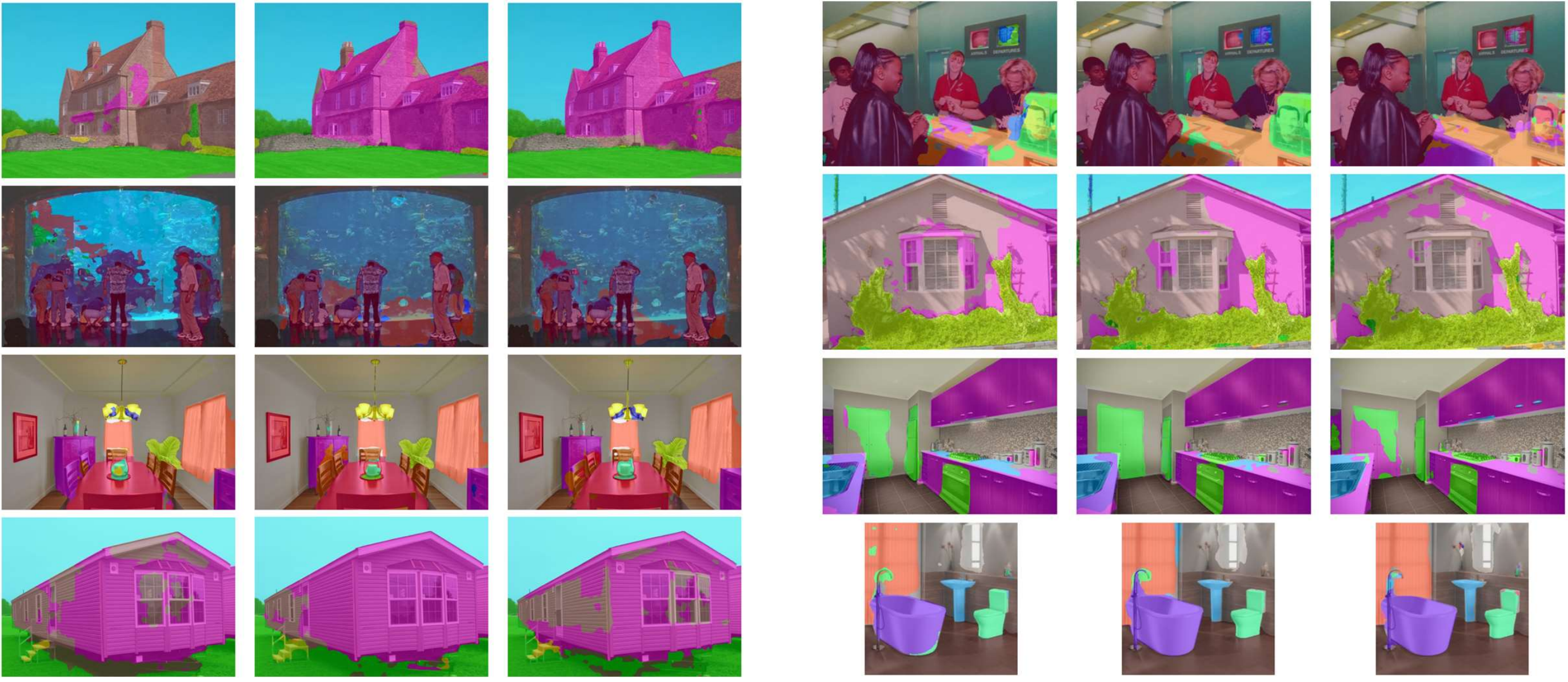}
    \caption{The visualization of the results of segmentation on ADE20K. From left to right, there is the original MAE, MAE with FAM, and MAE with FAMT, respectively.}
    \label{fig:seg_visual}
\end{figure*}

We performed fine-tuning on the \texttt{training} set for 80K iterations, and evaluated the models on the \texttt{validation} set. The results in Tab.~\ref{tab:segade} show that our method consistently outperforms the original MAE on all metrics. FAMT achieves the best performance on the $mIOU$ metric, which can be attributed to its ability to focus attention on fewer areas compared to attention-driven masking, enabling better object boundary detection. Additionally, both FAM and FAMT show improved performance on the $aAcc$ and $mAcc$ metrics compared to MAE with random masking.
Fig.~\ref{fig:seg_visual} displays the segmentation results on ADE20K, where FAM and FAMT outperform MAE with random masking. The segmentation of the human figure in the image is notably improved by FAMT, demonstrating its superior performance.

\subsubsection{Comparisons on iSAID} 
\begin{figure}[]
    \centering
    \includegraphics[width=0.5\textwidth]{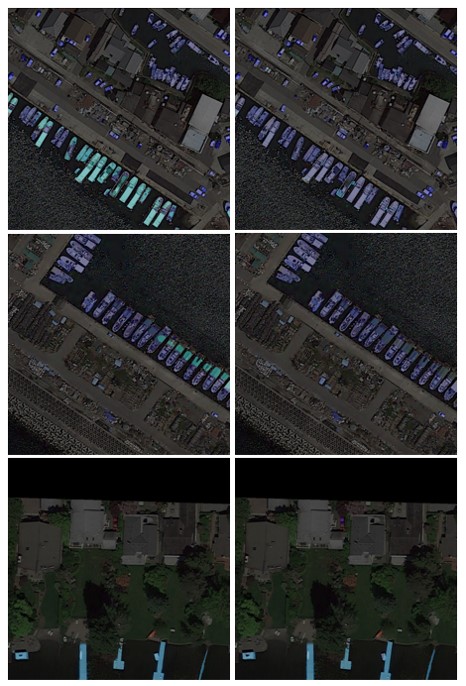}
    \caption{The visualization of the results of segmentation on iSAID. From left to right, there is the original MAE, and MAE with FAM, respectively.}
    \label{fig:isaid_visual}

\end{figure}

To investigate the transferability of our method, we report comparisons on iSAID. iSAID is a benchmark dataset of instance segmentation in autonomous driving scenarios, consisting of high-resolution images captured by UAV-mounted cameras in various urban and suburban areas of China. We fine-tune our models on the \texttt{train} set for 80K iterations and evaluate on the \texttt{val} set. As shown in Tab.~\ref{tab:segisaid}, our FAM method achieves a $1.1\%$ improvement in the $mIOU$ metric on iSAID, demonstrating its superiority. FAM also shows an obvious performance boost in terms of $aAcc$ and $mAcc$. However, compared to attention-driven masking, FAMT has a performance loss. This may be due to the small size of the objects in the images, which makes the throwing strategy of FAMT less effective in processing tiny objects. For each metric, both FAM and FAMT gain consistent performance improvements. The results are shown in Tab.~\ref{tab:segisaid}. Here we also provide the results of the segmentation in Fig.~\ref{fig:isaid_visual}. The ability of FAM to segment small objects is significantly better than original method.

\section{Ablations}
\label{sec:abl}

In this section, we design ablations from two aspects. One is FAMT, and the other is FAMT without frequency domain information, which we call AMT below.

   \begin{table*}[ht]
        \centering
        \small
        \renewcommand\arraystretch{1.2}
        \setlength\tabcolsep{3pt}
            \begin{tabular}{lcccccccccccc}
                \toprule
                \multicolumn{1}{c}{\multirow{2}{*}{Method}}  & \multicolumn{2}{c}{CIFAR-10}                    & \multicolumn{2}{c}{CIFAR-100}                   & \multicolumn{2}{c}{Tiny ImageNet}               & \multicolumn{2}{c}{STL-10}                      & \multicolumn{2}{c}{ImageNet} \\ \cmidrule(r){2-3}\cmidrule(r){4-5}\cmidrule(r){6-7}\cmidrule(r){8-9}\cmidrule(r){10-11}\cmidrule(r){12-13}
                \multicolumn{1}{c}{}                            & Linear                 & Fine-tuning            & Linear                 & Fine-tuning            & Linear                 & Fine-tuning            & Linear                 & Fine-tuning            & Linear     & Fine-tuning     \\ \midrule
                MAE+Random M                              & 85.2                   & 96.5                   & 65.2                   & 87.4                   & 55.2                   & 76.5                   & 80.9                   & 96.5                   & 56.6       & 82.6            \\
                MAE+AM (ours)              & \bf89.4 & 97.4                   & \bf69.9 & 87.3                   & \bf59.9 & 76.3                   & \bf87.1 & 97.4                   & 61.5       & 82.5            \\
                MAE+AMT (ours)                                    & 88.1                   & \bf97.5 & 68.7                   & \bf87.8 & 59.6                   & \bf77.8 & 86.8                   & \bf97.5 & \bf61.7       & \bf82.8            \\ \midrule
                SimMIM+Random M                                              & -                      & 95.0                   & -                      & 80.3                   & -                      & 74.0                   & -                      &  92.3                      & -           & \bf81.5            \\
                SimMIM+AM (ours)             & -                      & \bf97.8 & -                      & 85.9                   & -                      & \bf78.8 & -                      &          \bf96.5             &  -          & \bf81.5            \\
                SimMIM+AMT (ours)                                & -                      & 97.7                   & -                      & \bf86.1 & -                      & 75.8                   & -                      &    \bf96.5                  &   -         & 80.7            \\ 
                \bottomrule
                \end{tabular}
        \caption{Top-1 accuracy on CIFAR-10/100, Tiny ImageNet, STL-10, and ImageNet. M denotes Masking. AM denotes attention-driven masking. Random Masking is the default masking strategy for MAE and SimMIM.}
        \label{tab:classification400b}
        \end{table*}

\begin{table}[ht]
    \centering
    \small
    \renewcommand\arraystretch{1.2}
    \setlength\tabcolsep{3pt}
    \begin{tabular}{lcccr}
        \toprule
    \multicolumn{1}{c}{\multirow{2}{*}{SimMIM}} & \multicolumn{2}{c}{Ratio (\%)} & \multirow{2}{*}{\makecell[c]{Fine-tuning \\Top-1 Acc. (\%)}} & \multirow{2}{*}{\makecell[c]{Pre-training \\costs}} \\ 
    \multicolumn{1}{c}{} & Mask & \multicolumn{1}{l}{Throw} &  &  \\ \midrule
    +Random Masking & 60 & 0 &74.0  & 1.0$\times$ \\
    +AM & 60 & 0 & 78.8 & $\sim$1.0$\times$  \\
    +AMT & 44 & 26 & 75.8 &  $\sim$0.76$\times$\\
    +AMT & 33 & 50 & 75.1 & $\sim$0.62$\times$ \\
    \bottomrule
    \end{tabular}
    \caption{The accuracy on Tiny ImageNet and pre-training costs (i.e., time per epoch) using SimMIM with different ways of masking and throwing.}
    \label{tab:cost}
\end{table}

\subsection{AMT}

We conducted pre-training for MAE on ImageNet-1K for 400 epochs and 200 epochs respectively, using the same settings as the original works. The encoder backbone chosen was ViT-B/16, while the decoder for MAE was ViT-B/16 with 8 blocks, and for SimMIM, a linear head was used. In all experiments, absolute position embedding was used.

For linear probing, we trained a revised linear head with batch normalization and applied it for evaluation on classification with MAE. Fine-tuning was conducted using the same settings as the original works, with the \texttt{[CLS]} token used for both linear probing and fine-tuning.

For AMT, we use a throwing ratio of $t=0.4$ and $0.26$ for MAE and SimMIM, respectively. To maintain the ratio between masked and visible tokens, the masking ratio was set as $r=0.45$ and $0.44$ for MAE and SimMIM, respectively. This ensure that the impact of different ratios between masked and visible tokens was eliminated. The masking weights $a_w$ were updated every 40 epochs, which corresponded to 10\% of the whole pre-training process for MAE and 20\% for SimMIM. We use 4-GPU for this part.

Tab.~\ref{tab:classification400b} shows the evaluation results of the linear probing performance of MAE. Our AMT significantly improved the Top-1 accuracy by $2.9\% \sim 5.9\%$, indicating that our FAMT greatly enhanced the linear separability of the learned representations. Notably, the use of AMT only with 60\% of the images led to faster pre-training.

The quality of learned non-linear features can be assessed by the fine-tuning accuracy, which is crucial for downstream tasks. Tab.~\ref{tab:classification400b} presents the results of fine-tuning, where our AMT method outperforms the original random masking technique by $0.2\% \sim 5.8\%$. This improvement suggests that the representations learned by AMT possess higher transferability, which is a significant advantage for downstream tasks, the primary goal of self-supervised pre-training. Furthermore, our AMT method achieves comparable performance to attention-driven masking. Additionally, we observe that the selection of hyperparameters in our AMT is more extensive than in the original methods, leading to reduced time spent on searching for appropriate learning rates during fine-tuning.

As for detection and segmentation, to ensure a fair comparison, we adopt the same settings as ViTDet's detectron2 codebase~\cite{vitdet,wu2019detectron2}, using Mask R-CNN~\cite{Mask-rcnn} as the detector with a ViT-B/16 backbone and a total batch size of 16. Additionally, we keep all hyperparameters constant across all experiments.

\subsubsection{Comparisons on COCO} We carried out fine-tuning of our models on the \texttt{train2017} dataset for 90,000 iterations and evaluated the performance on the \texttt{val2017} set. The results in Tab.~\ref{tab:instance} indicate that our method consistently outperforms the original MAE in all metrics. Notably, our AMT method shows superior performance in most metrics, except for $AP^{mask}$, where it performs slightly worse than attention-driven masking due to its tendency to focus attention on fewer areas. However, AMT still performs well with a high threshold, and its more comprehensive attention to the object proves useful in segmentation tasks with a smaller threshold range.

\subsubsection{Comparisons on LVIS} To further examine the transferability of our method, we compare our results on the LVIS dataset. Unlike COCO, LVIS has imbalanced class distributions, and certain classes have less than 10 training examples. Additionally, the masks in LVIS are more concise and consistent, making detection and segmentation more challenging. We fine-tune our models on the \texttt{train} set for 75K iterations and evaluate on the \texttt{val} set. Tab.~\ref{tab:instance} demonstrates that our AMT method improves all metrics on this dataset by at least $1.4\%$. Our AMT still outperforms attention-driven masking, indicating the superiority of our approach.

\subsubsection{Different ways of masking and throwing}
We investigate the efficacy of different masking and throwing strategies in AMT pretraining, and also examine the impact of AMT on pretraining efficiency. To enhance representation learning using attention maps while reducing computational cost, we propose to discard certain tokens in our method. As displayed in Tab.~\ref{tab:ablation}, we report the classification accuracy results of six different masking and throwing schemes on Tiny ImageNet. The best performance is obtained by AMT with a $40\%$ throwing ratio and $45\%$ masking ratio, which enhances the baseline MAE by $1.0\%$. Notably, MAE with attention-driven masking only and discarding comparatively smaller fractions ($10\%$) of tokens both lead to accuracy decline. This observation indicates that attention-driven masking is slower to take effect than AMT. We further evaluate the effectiveness of attention-driven throwing by conducting experiments with random throwing. However, the random throwing approach performs poorly, highlighting the effectiveness of attention-driven throwing. Furthermore, we explore the impact of throwing different areas and find that discarding the medium-attended tokens, which is the default strategy of our AMT, outperforms discarding the low-attended tokens by $0.7\%$.

\subsubsection{Computing cost}
We conducted experiments on Tiny ImageNet to evaluate SimMIM using different masking and throwing techniques, and the outcomes are reported in Tab.~\ref{tab:cost}. All models underwent pre-training on the ImageNet-1K dataset for 200 epochs. Our attention-based masking and throwing strategy substantially boosted the performance while minimizing computational expenses. Notably, by discarding 50\% of the image, we achieved a 1.6$\times$ acceleration in pre-training time while maintaining better performance compared to the original SimMIM model that uses random masking.

\revisecolor{The acceleration effect of FAMT on pre-training stems from the throw operation, which selectively discards certain image tokens, completely excluding them from both the encoder's input and the decoder's reconstruction. Given that Transformer models exhibit a sample complexity of \(O(n^2)\), the reduction in the number of tokens leads to a quadratic saving in computational expense.}

\subsubsection{Different settings for hypermeters}
\revisecolor{In Table \ref{tab:hyper}, we concurrently explored the impact of different settings for the hyperparameters 
throw ratio $t$ and mask ration $r$ on the experimental outcomes. The results indicate that discarding a proportion of image patches ranging from 40\% to 70\% significantly enhances the linear probing performance of our method. Notably, the best experimental results were achieved at discard ratios of 20\% and 70\%. It can be observed that we maintained a ratio of 1:3 between visible patches and mask patches to eliminate the effects arising from variations in this ratio. Additionally, the default setting in this paper is to discard 40\% of the image patches.}
\begin{table}[t]
\centering
    \small
    \renewcommand\arraystretch{1.2}
\begin{tabular}{ccccc}
\toprule
\multirow{2}{*}{Method}   & \multicolumn{3}{c}{Ratio(\%)} & \multirow{2}{*}{Acc.} \\ \cmidrule{2-4}
                          & mask    & throw   & visible   &                       \\ \midrule
\multirow{10}{*}{MAE+AMT} & 7.5     & 90      & 2.5       & 27.9                  \\
                          & 15      & 80      & 5         & 31.6                  \\
                          & 22.5    & 70      & 7.5       & \underline{33.3}                  \\
                          & 30      & 60      & 10        & \bf{33.8}                  \\
                          & 37.5    & 50      & 12.5      & 33.0                  \\
                          & 45      & 40      & 15        & 32.7                  \\
                          & 52.5    & 30      & 17.5      & 32.1                  \\
                          & 60      & 20      & 20        & \bf{33.8}                  \\
                          & 67.5    & 10      & 22.5      & 31.4                  \\
                          & 75      & 0       & 25        & 31.3                  \\ \midrule
MAE+random masking        & 75      & 0       & 25        & 32.2  \\  \bottomrule             
\end{tabular}
\caption{Top-1 linear probing accuracy on Tiny-ImageNet. All models are pretrained on Tiny-ImageNet for 200 epochs.}
\label{tab:hyper}
\end{table}
\subsection{FAMT}
As shown in Tab.~\ref{tab:ablation pretrain on tiny}, we design 5 different strategies with MAE using ViT-S/16. All the models are pretrained on Tiny ImageNet for 400 epochs. The linear probing accuracy and finetuning accuracy are listed in the table. We can clearly find that MAE with FAMT has the best accuracy, which significantly outperforms other strategies and achieves a 32.9\% classification accuracy.

\revisecolor{Comparing the results of all ablation experiments, it can be found that the throwing operation can largely improve the accuracy of linear probing, but will reduce the accuracy of finetuning. That is also why MAE with FAM has the best finetuning accuracy.} When using attention-driven masking without frequency domain information, the linear accuracy has been affected. We think it is because too much attention to spatial domain information can diminish the linear classification performance of ViT to some extent. The introduction of frequency domain information at this time can better utilize the low-pass performance of ViT. This is also confirmed by the experimental results in the bottom two rows of Tab.~\ref{tab:ablation pretrain on tiny}. Masking and throwing based on frequency domain information show good affinity, improving the linear classification ability of the model together.

\section{Conclusion}
FAMT employs the self-attention mechanism in ViT for masking and throwing parts of the input image. By utilizing the semantic information learned by the model during training, FAMT can help the model focus on the object and ignore the background, leading to improved performance and reduced computational cost. In addition, FAMT incorporates frequency domain information for token selection, enabling it to leverage ViT's low-pass filtering ability. FAMT is a modular plug-and-play component for masked image modeling that can be easily integrated into MIM methods that use ViT as their backbone. We chose MAE and SimMIM because they are the most typical and pure MIM models. \revisecolor{After demonstrating the effectiveness of FAMT on them, any MIM method can easily benefit from the gains provided by FAMT.} Our experiments show that incorporating FAMT into typical MIM methods such as MAE and SimMIM results in superior performance on a range of downstream datasets, demonstrating the transferability of the learned representations. We hope that our work will inspire further research in this area.


  


\bibliographystyle{IEEEtran}
\bibliography{IEEEabrv,references}

 \begin{IEEEbiography}[{\includegraphics[width=1in,height=1.25in,clip,keepaspectratio]{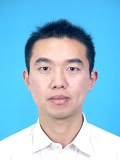}}]{Jie Gui} (SM'16) is currently a professor at the School of Cyber Science and Engineering, Southeast University. He received a BS degree in Computer Science from Hohai University, Nanjing, China, in 2004, an MS degree in Computer Applied Technology from the Hefei Institutes of Physical Science, Chinese Academy of Sciences, Hefei, China, in 2007, and a PhD degree in Pattern Recognition and Intelligent Systems from the University of Science and Technology of China, Hefei, China, in 2010. He has published more than 60 papers in international journals and conferences such as IEEE TPAMI, IEEE TNNLS, IEEE TCYB, IEEE TIP, IEEE TCSVT, IEEE TSMCS, KDD, and ACM MM. He is the Area Chair, Senior PC Member, or PC Member of many conferences such as NeurIPS and ICML. He is an Associate Editor of IEEE Transactions on Circuits and Systems for Video Technology (T-CSVT), Artificial Intelligence Review, Neural Networks, and Neurocomputing. His research interests include machine learning, pattern recognition, and image processing.
 \end{IEEEbiography}

\begin{IEEEbiography}[{\includegraphics[width=1in,height=1.25in,clip,keepaspectratio]{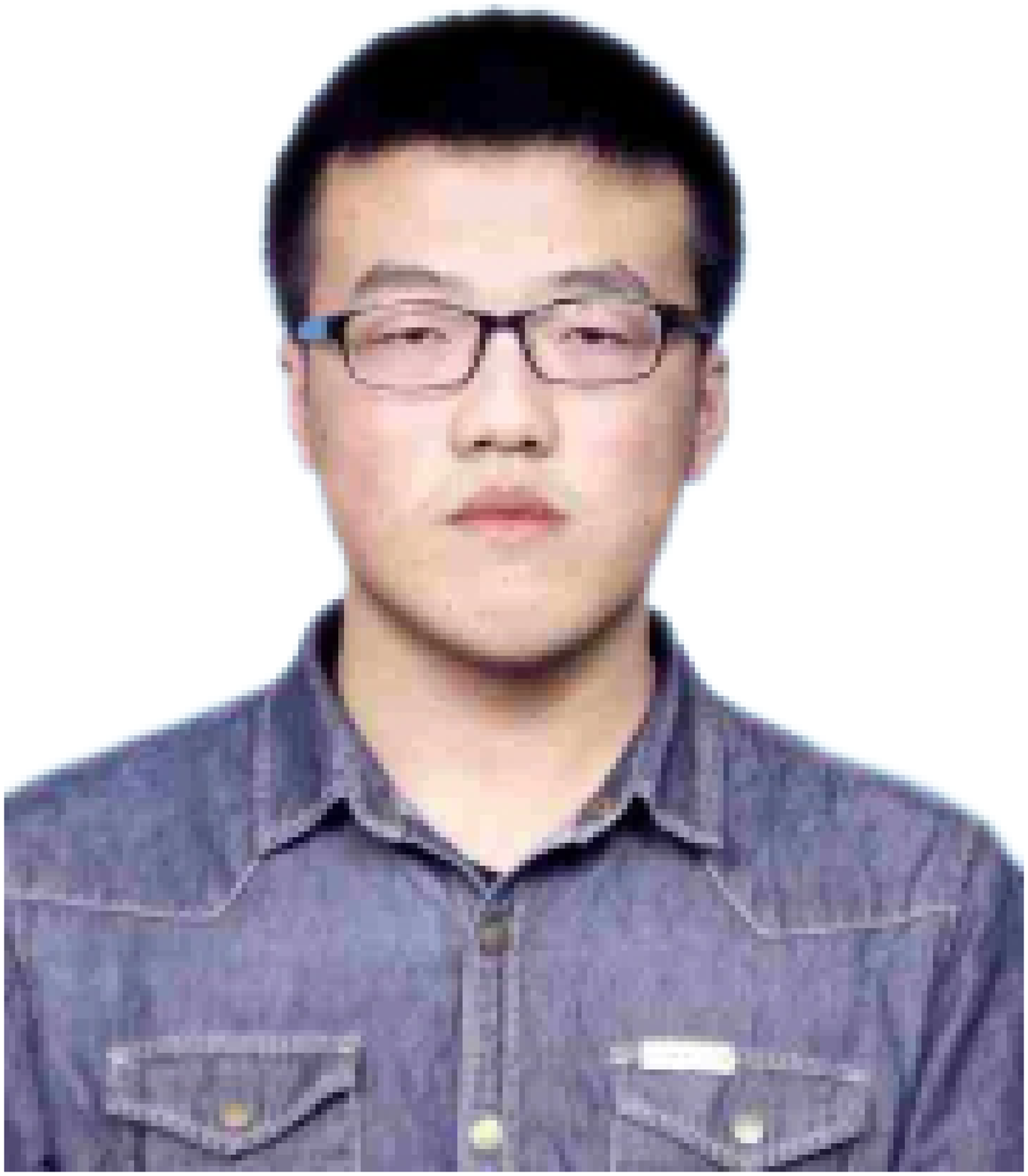}}]{Tuo Chen} is a PhD student with
 the Department of Electronic Information, Southeast University. He received his bachelor’s degree from the Department of Information Security, Lanzhou University. His main research interests include self-supervised learning and adversarial examples.
 \end{IEEEbiography}

\begin{IEEEbiography}[{\includegraphics[width=1in,height=1.25in,clip,keepaspectratio]{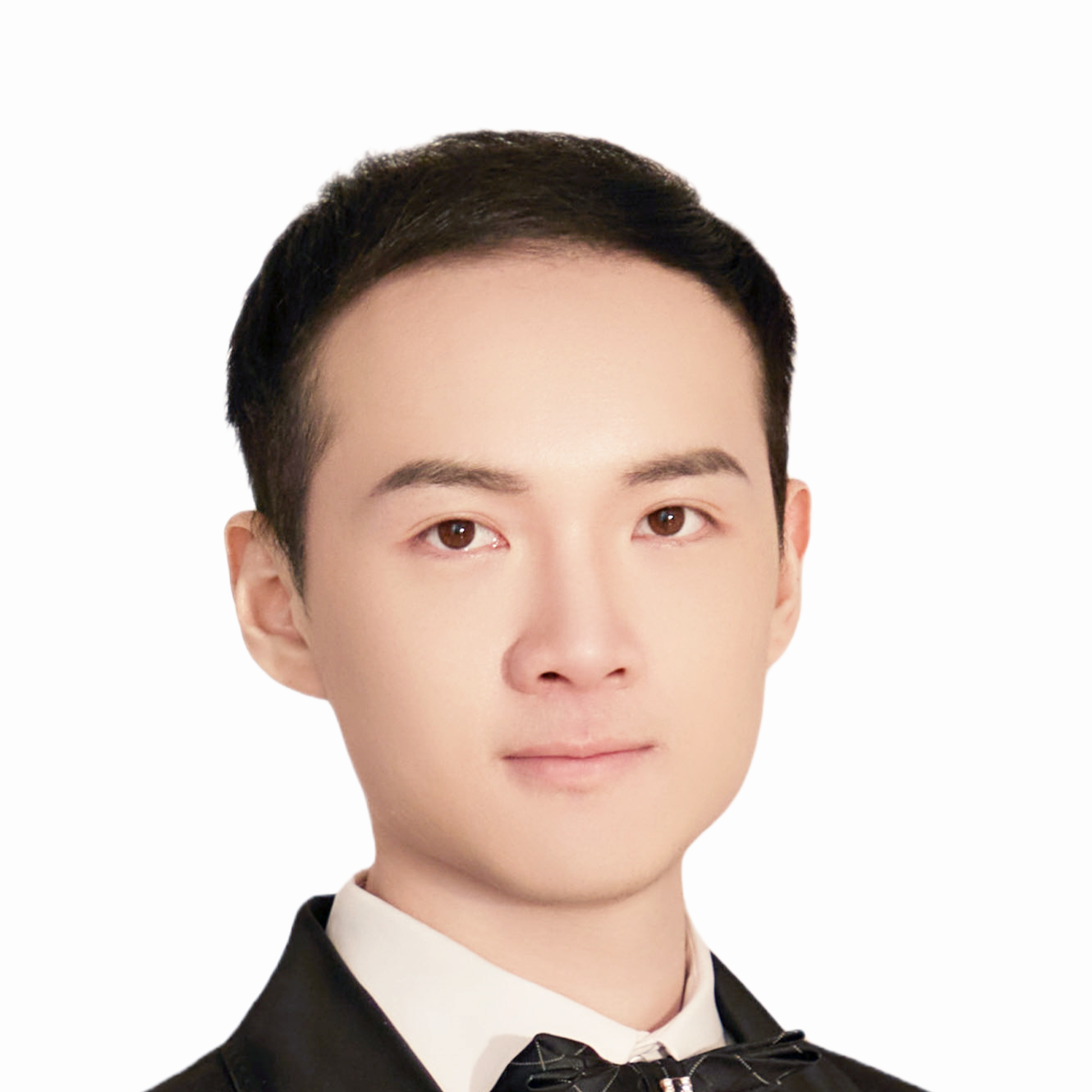}}]{Minjing Dong} is an Assistant Professor at the Department of Computer Science, City University of Hong Kong since January 2024. He received his Ph.D. and M.Phil degree from School of Computer Science, University of Sydney, supervised by Dr. Chang Xu.
 \end{IEEEbiography}
 
  \begin{IEEEbiography}[{\includegraphics[width=1in,height=1.25in,clip,keepaspectratio]{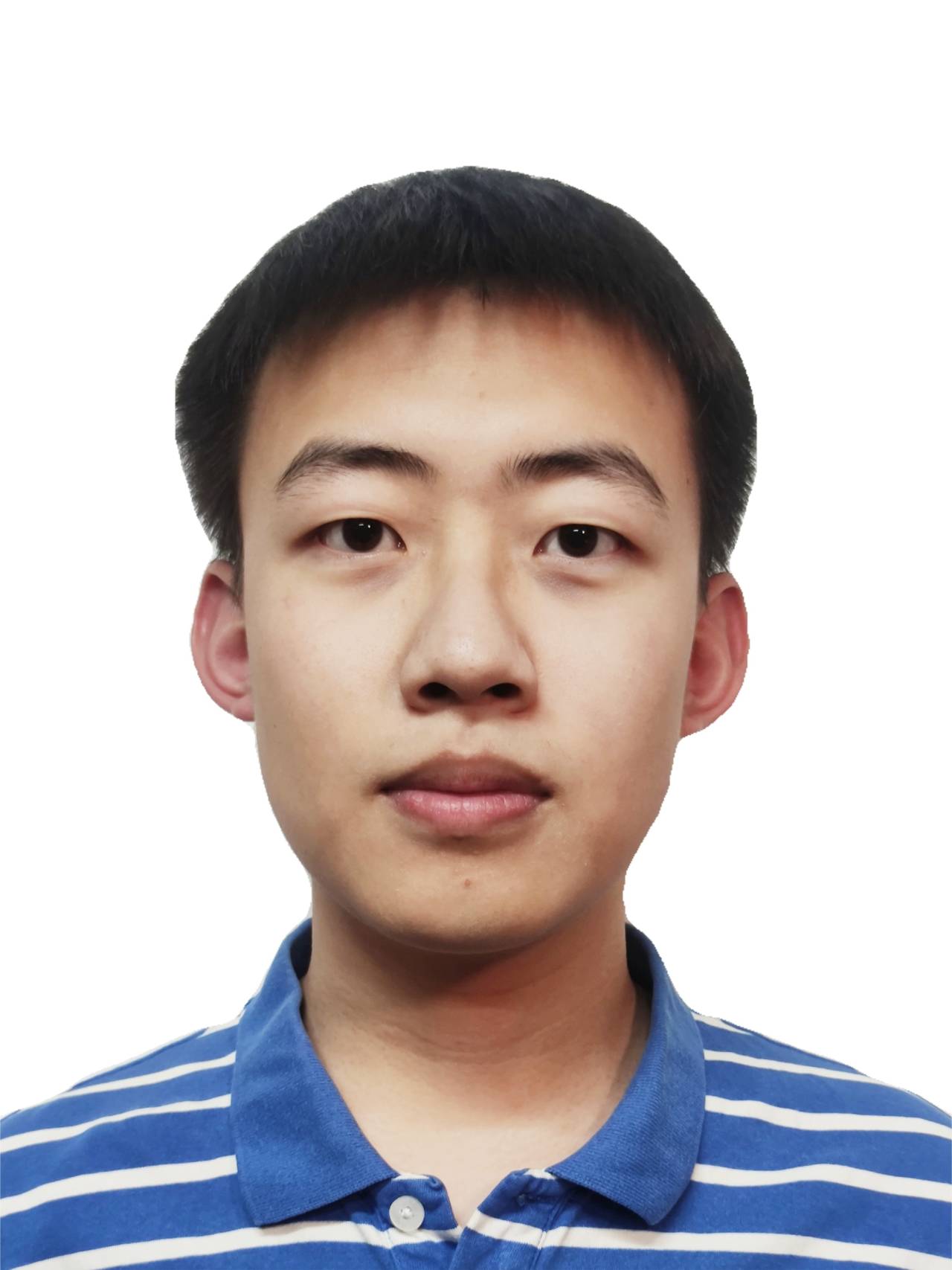}}]{Zhengqi Liu} is now working toward the M.S. degree from the School of Cyber Science and Engineering, Southeast, University. He received his bachelor’s degree from the School of Automation, Southeast University. His main research interests include self-supervised learning.
 \end{IEEEbiography}

 \begin{IEEEbiography}[{\includegraphics[width=1in,height=1.25in,clip,keepaspectratio]{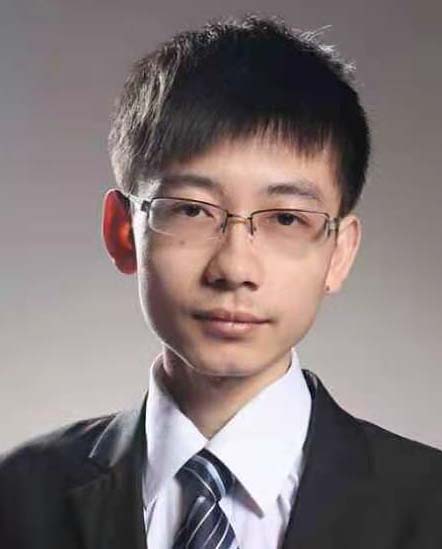}}]{Hao Luo} received B.S. and PhD degrees from Zhejiang University, China, in 2015 and 2020, respectively. He is currently working at the Alibaba DAMO Academy. His research interests include person re-identification, vision transformer, self-supervised, computer vision, and deep learning.
\end{IEEEbiography}

  \begin{IEEEbiography}[{\includegraphics[width=1in,height=1.25in,clip,keepaspectratio]{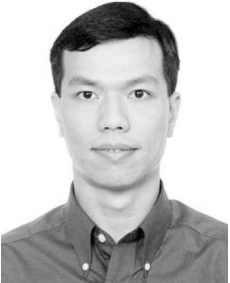}}]{James Tin-Yau Kwok} (Fellow, IEEE) received the Ph.D. degree in computer science from The Hong Kong University of Science and Technology, Hong Kong, in 1996. He is currently a Professor with the Department of Computer Science and Engineering, The Hong Kong University of Science and Technology. His current research interests include kernel methods, machine learning, pattern recognition, and artificial neural networks. He received the IEEE Outstanding Paper Award in 2004 and the Second Class Award in Natural Sciences from the Ministry of Education, China, in 2008. He has been a Program Co-Chair for a number of international conferences, and served as an Associate Editor for the IEEE TRANS-ACTIONS ON NEURAL NETWORKS AND LEARNING SYSTEMS from 2006 to 2012. He is currently an Associate Editor of Neurocomputing.
\end{IEEEbiography}

  \begin{IEEEbiography}[{\includegraphics[width=1in,height=1.25in,clip,keepaspectratio]{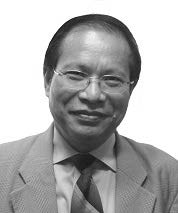}}]{Yuan Yan Tang} (F'04) is an IEEE Life Fellow, IAPR Fellow, and AAIA Fellow. He currently is the Director of Smart City Research Center in Zhuhai UM Science \& Technology Research Institute, is also the Emeritus Chair Professor at University of Macau and Hong Kong Baptist University, Adjunct Professor at Concordia University, Canada. His current research interests include artificial intelligence, wavelets, pattern recognition, and image processing. He has published more than 600 academic papers and is the author (or coauthor) of over 25 monographs, books and bookchapters. He is the Founder and Editor-in-Chief of SCI journal ``International Journal on Wavelets, Multiresolution, and Information Processing (IJWMIP)''. Dr. Tang is the Founder and General Chair of the series International Conferences on Wavelets Analysis and Pattern Recognition (ICWAPRs). He is the Founder and Chair of the Macau Branch of International Associate of Pattern Recognition (IAPR). He has serviced as general chair, program chair, and committee member for many international conferences. Dr. Tang served as the Chairman of 18th ICPR, which is the first time that the ICPR was hosted in China.
\end{IEEEbiography}

\end{document}